\documentclass[runningheads]{llncs}
\usepackage{graphicx}
\usepackage{amssymb}

\begin{document}
\title{Siamese Survival Analysis with Competing Risks}
\author{Anton Nemchenko\inst{1} \and
Trent Kyono\inst{1} \and
Mihaela Van Der Schaar\inst{1,2,3}}
\authorrunning{A. Nemchenko et al.}
\institute{University of California, Los Angeles, Los Angeles CA 90095, USA \and 
University of Oxford, Oxford OX1 2JD, UK \and
Alan Turing Institute, 96 Euston Rd, Kings Cross, London NW1 2DB, UK}
\maketitle              
\begin{abstract}
Survival analysis in the presence of multiple possible adverse events, i.e., competing risks, is a pervasive problem in many industries (healthcare, finance, etc.).  Since only one event is typically observed, the incidence of an event of interest is often obscured by other related competing events. 
This nonidentifiability, or inability to estimate true cause-specific survival curves from empirical data, further complicates competing risk survival analysis. We introduce Siamese Survival Prognosis Network (SSPN),  a novel deep learning architecture for estimating personalized risk scores in the presence of competing risks. SSPN circumvents the nonidentifiability problem by avoiding the estimation of cause-specific survival curves and instead determines pairwise concordant time-dependent risks, where longer event times are assigned lower risks. Furthermore, SSPN is able to directly optimize an approximation to the C-discrimination index, rather than relying on well-known metrics which are unable to capture the unique requirements of survival analysis with competing risks.

\end{abstract}
\section{Introduction}
\subsection{Motivation}
Survival analysis is a method for analyzing data where the outcome variable is the time to the occurrence of an event (death, disease, stock liquidation, mechanical failure, etc.) of interest.  Competing risks are additional possible events or outcomes that ``compete" with and may preclude or interfere with the desired event observation.  Though survival analysis is practiced across many disciplines (epidemiology, econometrics, manufacturing, etc.), this paper focuses on healthcare applications, where competing risk analysis has recently emerged as an important analytical tool in medical prognosis \cite{glynn2005comparison,wolbers2009prognostic,satagopan2004note}.  With an increasing aging population, the presence of multiple coexisting chronic diseases (multimorbidities) is on the rise, with more than two-thirds of people aged over 65 considered multimorbid.  Developing optimal treatment plans for these patients with multimorbidities is a challenging problem, where the best treatment or intervention for a patient may depend upon the existence and susceptibility to other competing risks.  Consider oncology and cardiovascular medicine, where the risk of a cardiac disease may alter the decision on whether a cancer patient should undergo chemotherapy or surgery.  Countless examples like this involving competing risks are pervasive throughout the healthcare industry and insufficiently addressed in it's current state.
\subsection{Related Works}
Previous work on classical survival analysis has demonstrated the advantages of deep learning over statistical methods \cite{luck2017deep,katzman2016deep,yousefi2017predicting}. Cox proportional hazards model \cite{cox1972models} is the baseline statistical model for survival analysis, but is limited since the dependent risk function is the product of a linear covariate function and a time dependent function, which is insufficient for modeling complex non-linear medical data.
\cite{katzman2016deep} replaced the linear covariate function with a feed-forward neural network as input for the Cox PH model and demonstrated improved performance. The current literature addresses competing risks based on statistical methods (the Fine Gray model \cite{fine1999proportional}), classical machine learning (Random Survival Forest \cite{ishwaran2008random,ishwaran2014random}), multi-task learning \cite{alaadeep}) etc., with limited success. These existing competing risk models are challenged by computational scalability issues for datasets with many patients and multiple covariates.  To address this challenge, we propose a deep learning architecture for survival analysis with competing risks to optimize the time-dependent discrimination index.  This is not trivial and will be elaborated in the next section.
\subsection{Contributions}
In both machine learning and statistics, predictive models are compared in terms of the area under the receiver operating characteristic (ROC) curve or the time-dependent discrimination index (in the survival analysis literature). The equivalence of the two metrics was established in \cite{heagerty2005survival}. Numerous works on supervised learning \cite{chen2013gradient,mayr2014boosting,mayr2016boosting,schmid2016use} have shown that training the models to directly optimize the AUC improves out-of-sample (generalization) performance (in terms of AUC) rather than optimizing the error rate (or the accuracy). In this work, we adopt and apply this idea to survival analysis with competing risks. We develop a novel Siamese feed-forward neural network \cite{bromley1994signature} designed to optimize concordance and account for competing risks by specifically targeting the time-dependent discrimination index \cite{antolini2005time}. This is achieved by estimating risks in a relative fashion so that the risk for the ``true'' event of a patient (i.e. the event which actually took place) must be higher than: all other risks for the same patient and the risks for the same true event of other patients that experienced it at a later time. Furthermore, the risks for all the causes are estimated jointly in an effort to generate a unified representation capturing the latent structure of the data and estimating cause-specific risks. Because our neural network issues a joint risk for all competing events, it compares different risks for the different events at different times and arranges them in a concordant fashion (earlier time means higher risk for any pair of patients).

Unlike previous Siamese neural networks architectures \cite{chopra2005learning,bromley1994signature,wang2017multi} developed for purposes such as learning the pairwise similarity between different inputs, our architecture aims to maximize the distance between output risks for the different inputs. We overcome the discontinuity problem of the above metric by introducing a continuous approximation of the time-dependent discrimination function. This approximation is only evaluated at the survival times observed in the dataset. However, training a neural network only over the observed survival times will result in poor generalization and undesirable out-of-sample performance (in terms of discrimination index computed at different times). In response to this, we add a loss term (to the loss function) which for any pair of patients, penalizes cases where the longer event time does not receive lower risk. 

The nonidentifiability problem in competing risks arises from the inability to estimate the true cause-specific survival curves from empirical data \cite{tsiatis1975nonidentifiability}.  We address this issue by bypassing and avoiding the estimation of the individual cause-specific survival curves and utilize concordant risks instead.  Our implementation is agnostic to any underlying causal assumptions and therefore immune to nonidentifiability.

We report statistically significant improvements over state-of-the-art competing risk survival analysis methods on both synthetic and real medical data.

\section{Problem Formulation}
We consider a dataset $\mathcal{H}$ comprising of time-to-event information about $N$ subjects who are followed up for a finite amount of time.  Each subject (patient) experiences an event $D\in \{0,1,..,M\}$, where $D$ is the event type. $D=0$ means the subject is censored (lost in follow-up or study ended). If $D\in \{1,..,M\}$, then the subject experiences one of the events of interest (for instance, subject develops cardiac disease). We assume that a subject can only experience one of the above events and that the censorship times are independent of them \cite{lambert2010estimating,satagopan2004note,fine1999proportional,crowder2001classical,gooley1999estimation,tsiatis1975nonidentifiability}. $T$ is defined as the time-to-event, where we assume that time is discrete $T\in \{t_1,...,t_K\}$ and $t_1=0$ ($t_i$ denotes the elapsed time since $t_1$). Let $\mathcal{H} = \{T_i, D_i, x_i\}_{i=1}^{N}$, where $T_i$ is the time-to-event for subject $i$,  $D_i$ is the event experienced by the subject $i$ and $x_i \in \mathbb{R}^{S}$ are the covariates of the subject (the covariates are measured at baseline, which may include age, gender, genetic information etc.).

The Cumulative Incidence Function (CIF) \cite{fine1999proportional} computed at time $t$ for a certain event $D$ is the probability of occurrence of a particular event $D$ before time $t$ conditioned on the covariates of a subject $x$, and is given as $F(t,D|x) = Pr(T\leq t, D| x)$. The cumulative incidence function evaluated at a certain point can be understood as the risk of experiencing a certain event before a specified time.

In this work, our goal is to develop a neural network that can learn the complex interactions in the data specifically addressing competing risks survival analysis.  In determining our loss function, we consider that the time-dependent discrimination index is the most commonly used metric for evaluating models in survival analysis \cite{antolini2005time}.   
Multiple publications in the supervised learning literature demonstrate that approximating the area under the curve (AUC) directly and training a classifier leads to better generalization performance in terms of the AUC (see e.g. \cite{chen2013gradient,mayr2014boosting,mayr2016boosting,schmid2016use}).  However, these ideas were not explored in the context of survival analysis with competing risks. We will follow the same principles to construct an approximation of the time-dependent discrimination index to train our neural network. We first describe the time-dependent discrimination index below.

Consider an ordered pair of two subjects $(i,j)$ in the dataset. If the subject $i$ experiences event $m$, i.e., $D_i \not= 0$ and if subject $j$'s time-to-event exceeds the time-to-event of subject $i$, i.e., $T_j > T_i$, then the pair $(i,j)$ is a comparable pair. The set of all such comparable pairs is defined as the comparable set for event $m$, and is denoted as $X^m$. 

A model outputs the risk of the subject $x$ for experiencing the event $m$ before time $t$, which is given as $R^m(t,x)=F(t,D=m|x)$. The time-dependent discrimination index for a certain cause $m$ is the probability that a model accurately orders the risks of the comparable pairs of subjects in the comparable set for event $m$.  The time-dependent discrimination index \cite{antolini2005time} for cause $m$ is defined as   
\begin{equation} \label{eq:eq1}
C_{\mbox{t}}(m)=\frac{\sum_{k=1}^KAUC^m(t_k)w^m(t_k)}{\sum_{k=1}^Kw^m(t_k)}\ .
\end{equation}
where
\begin{equation}
AUC^m(t_k)=Pr\{R^m(t_k,x_i)>R^m(t_k,x_j)|T_i=t_k,T_j>t_k,D_i=m\}\ ,
\end{equation}
\begin{equation}
w^m(t_k)=Pr\{T_i=t_k,T_j>t_k,D_i=m\}\ .
\end{equation}
The discrimination index in (\ref{eq:eq1}) cannot be computed exactly since the distribution that generates the data is unknown. However, the discrimination index can be estimated using a standard estimator, which takes as input the risk values associated with subjects in the dataset. \cite{antolini2005time} defines the estimator for (\ref{eq:eq1}) as 
\begin{equation} \label{eq:eq2}
\hat{C}_{\mbox{t}}(m)=\frac{\sum_{i,j=1}^{N}\textbf{1}\{R^m(T_i,x_i)>R^m(T_i,x_j)\}\cdot \textbf{1}\{T_j>T_i,D_i=m\}}
{\sum_{i,j=1}^{N}\textbf{1}\{T_j>T_i,D_i=m\}}\ . 
\end{equation}
Note that in the above (\ref{eq:eq2}) only the numerator depends on the model. Henceforth, we will only consider the quantity in the numerator and we write it as
\begin{equation}
\bar{C}_{\mbox{t}}(m)=\sum_{i,j=1}^{N}\textbf{1}\{R^m(T_i,x_i)>R^m(T_i,x_j)\}\cdot \textbf{1}\{T_j>T_i,D_i=m\}\ .
\end{equation}
The above equation can be simplified as 
\begin{equation} \label{eq:eq3}
\bar{C}_{\mbox{t}}(m)=\sum_{i=1}^{|X^m|}\textbf{1}\{R^m(T_i(\mbox{left}),  X^m_i(\mbox{left}))>R^m(T_i(\mbox{left}),X^m_i(\mbox{right}))\}\ .
\end{equation}
where $\textbf{1}(x)$ is the indicator function, $X^m_i(\mbox{left})$ ($X^m_i(\mbox{right})$) is the left (right) element of the $i^{th}$ comparable pair in the set $X^{m}$ and $T_i(\mbox{left})$ ($T_i(\mbox{right})$) is the respective time-to-event. In the next section, we will use the above simplification (\ref{eq:eq3}) to construct the loss function for  the neural network.
\begin{figure}
  \centering
  \includegraphics[width=0.52\textwidth]{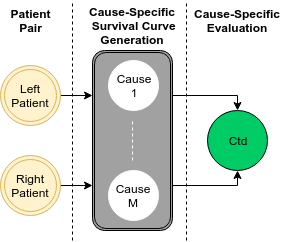}
  \caption{Illustration of the architecture.}
  \label{fig:fig1}
\end{figure}
\section{Siamese Survival Prognosis Network}
In this section, we will describe the architecture of the network and the loss functions that we propose to train the network.

Denote $H$ as a feed-forward neural network which is visualized in Fig. \ref{fig:fig1}. It is composed of a sequence of $L$ fully connected hidden layers with ``scaled exponential linear units'' (SELU) activation. The last hidden layer is fed to $M$ layers of width $K$. Each neuron in the latter $M$ layers estimates the probability that a subject $x$ experiences cause $m$ occurs in a time interval $t_k$, which is given as $Pr^{m}(t_k,x)$.  For an input covariate $x$ the output from all the neurons is a vector of probabilities given as $ \Big\{\big[Pr^{m}(t_k,x)\big]_{k=1}^{K}\Big\}_{m=1}^{M}$. \\
The estimate of cumulative incidence function computed for cause $m$ at time $t_k$ is given as $\tilde{R}^m(t_k,x)=\sum_{i=1}^{k}Pr^m(t_i,x)$. The final output of the neural network for input $x$ is vector of estimates of the cumulative incidence function given as $H(x) =  \Big\{\big[\tilde{R}^{m}(t_k,x)\big]_{k=1}^{K}\Big\}_{m=1}^{M}$. 

The loss function is composed of three terms: discrimination, accuracy, and a loss term. 

We cannot use the metric in (\ref{eq:eq3})  directly to train the network because it is a discontinuous function (composed of indicators), which can impede training. We overcome this problem by approximating the indicator function using a scaled sigmoid function $\sigma(\alpha x) = \frac{1}{1+exp(-\alpha x)}$. The approximated discrimination index is given as 
\begin{equation}
\hat{\bar{C}}_{\mbox{t}}(m)=\sum_{i=1}^{|X^m|}\sigma\Big[\alpha\big[\tilde{R}^m(T_i(\mbox{left}),X^m_i(\mbox{left}))-\tilde{R}^m(T_i(\mbox{left}),X^m_i(\mbox{right}))\big]\Big]\ .
\end{equation}
The scaling parameter $\alpha$ determines the sensitivity of the loss function to discrimination. If the value of $\alpha$ is high, then the penalty for error in discrimination is also very high. Therefore, higher values of alpha guarantee that the subjects in a comparable pair are assigned concordant risk values.  

The discrimination part defined above captures a model's ability to discriminate subjects for each cause separately. We also need to ensure that the model can predict the cause accurately. We define the accuracy of a model in terms of a scaled sigmoid function with scaling parameter $\kappa$ as follows
\begin{equation}
L^1=\sum_{i=1}^{|X^m|}\sigma\Big[\kappa \big(\tilde{R}^{D(\mbox{left})}(T_i(\mbox{left}),X^m_i(\mbox{left}))-\sum_{m \neq D(\mbox{left})}\tilde{R}^m(T_i(\mbox{left}),X^m_i(\mbox{left}))\big)\Big]\ .
\end{equation}
The accuracy term penalizes the risk functions only at the event times of the left subjects in comparable pairs. However, it is important that the neural network is optimized to produce risk values that interpolate well to other time intervals as well. Therefore, we introduce a loss term below 
\begin{equation}
L^2=\beta\sum_{m=1}^{M} \sum_{i=1}^{|X^{m}|}\sum_{t_k<T_i(\mbox{left})}R^m(t_k,X^m_i(\mbox{right}))^2\ .
\end{equation}
The loss term ensures that the risk of each right subject is minimized for all the times before  time-to-event of the left subject in the respective comparable pair. Intuitively, the loss term can be justified as follows.  The right subjects do not experience an event before the time $T_i(\mbox{left})$. Hence, the probability that they experience an event before $T_i(\mbox{left})$ should take a small value. 

The final loss function is the sum of the discrimination terms (described above),  the accuracy and the loss terms, and is given as
\begin{equation} \label{eq:eq4}
\sum_{m=1}^{M}\hat{\bar{C}}_{\mbox{t}}(m) + L^1 + L^2\ . 
\end{equation} 
Finally, we adjust for the event imbalance and the time interval imbalance caused by the unequal number of pairs for each event and time interval with inverse propensity weights. These weights are the frequency of the occurrence of the various events at the various times and are multiplying the loss functions of the corresponding comparable pairs. 

We train the feed-forward network using the above loss function (\ref{eq:eq4}) and regularize it using SELU dropout \cite{klambauer2017self}. Since the loss function involves the discrimination term, each term in the loss function involves a pairwise comparison. This makes the network training similar to a Siamese network \cite{bromley1994signature}. The backpropagation terms now depend on each comparable pair.  
\section{Experiments}
This section includes a discussion of hyper-parameter optimization followed by competing risk and survival analysis experiments\footnote{Code available at \url{https://github.com/santon834/Siamese-Competing-Risks}}. We compare against Fine-Gray model (``cmprsk" R package), Competing Random Forest (CRF) (``randomForestSRC" R package) and the cause-specific (cs) extension of two single event (non-competing risks) methods, Cox PH model and \cite{katzman2016deep}. In cause-specific extension of single event models, we mark the occurrence of any event apart from the event of interest as censorship and decouple the problem into separate single event problem (one for each cause); this is a standard way of extending single-event models to competing risk models. In the following results we refer to our method with the acronym SSPN.
\begin{table}
  \caption{Summary of hyper-parameters}
  \label{tbl:tbl1}
  \centering
  \begin{tabular}{ l | c | c | c | c }
	\hline
	\noalign{\smallskip}                 
    Parameter	& batch size & \# hidden layers & hidden layers width & dropout rate \\
    \noalign{\smallskip}
	\hline
	\noalign{\smallskip}
    SEER & 2048 & 3 & 50 & 0.4 \\
    Synthetic data & 2048 & 2 & 40 & 0.35 \\
    \noalign{\smallskip}
	\hline
	\noalign{\smallskip}
  \end{tabular}
\end{table}
\subsection{Hyper-parameter Optimization}
Optimization was performed using a 5-fold cross-validation with fixed censorship rates in each fold. We choose 60-20-20 division for training, validation and testing sets. A standard grid search was used to determine the batch size, number of hidden layers, width of the hidden layers and the dropout rate. The optimal values of $\alpha$ and $\beta$ were consistently 500 and 0.01 for all datasets.  As previously mentioned, the sets are comprised of patient pairs. In each training iteration, a batch size of pairs was sampled with replacement from the training set which reduces convergence speed but doesn't lower performance relative to regular batches \cite{Recht}. We note that the training sets are commonly in the tens of million pairs with patients appearing multiple times in both sides of the pair. A standard definition of an epoch would compose of a single iteration over all patient. However, in our case, we not only learn patient specific characteristics but also patient comparison relationships, which means an epoch with a number of iterations equal to the number of patients is not sufficient. On the other hand, an epoch definition as an iteration over all pairs is impractical. Our best empirical results were attained after 100K iterations with Tensorflow on 8-core Xeon E3-1240, Adam optimizer \cite{kingma2014adam} and a decaying learning rate, $LR^{-1}(i)=10^{-3}+i$. Table \ref{tbl:tbl1} summarizes the optimal hyper-parameters.
\begin{table}
  \caption{Summary of competing $C_{\mbox{t}}$ index on SEER.}
  \label{tbl:tbl2}
  \centering
  \begin{tabular}{ l | c | c | c }
	\hline
	\noalign{\smallskip}                 
    Dataset	& CVD & Breast Cancer & Other \\
    \noalign{\smallskip}
	\hline
	\noalign{\smallskip}
    cs-Cox PH & 0.656 [0.629-0.682] & 0.634 [0.626-0.642] &  0.695 [0.675-0.714] \\
    cs-\cite{katzman2016deep} & 0.645 [0.625-0.664] & 0.697 [0.686-0.708] & 0.675 [0.644-0.706]	\\
    Fine-Gray & 0.659 [0.605-0.714] & 0.636 [0.622-0.650] & 0.691 [0.673-0.708] \\
    CRF & 0.601 [0.565-0.637] & 0.705 [0.692-0.718] & 0.636 [0.624-0.648] \\
    \noalign{\smallskip}
	\hline
	\noalign{\smallskip}
    SSPN 	& \textbf{0.663 [0.625-0.701]} & \textbf{0.735 [0.678-0.793]} & \textbf{0.699 [0.681-0.716]} \\
    \noalign{\smallskip}
	\hline
	\noalign{\smallskip}
    \multicolumn{4}{l}{*p-value $<$ 0.05}
  \end{tabular}
  \caption{Summary of competing $C_{\mbox{t}}$ index on synthetic data.}
  \label{tbl:tbl3}
  \centering
  \begin{tabular}{ l | c | c }
	\hline
	\noalign{\smallskip}                 
    Method	& Cause 1 & Cause 2 \\
    \noalign{\smallskip}
	\hline
	\noalign{\smallskip}
    cs-Cox PH & 0.571 [0.554-0.588] & 0.581 [0.570-0.591] \\
    cs-\cite{katzman2016deep} & 0.580 [0.556-0.603] & 0.593 [0.576-0.611]  \\
    Fine-Gray & 0.574 [0.559-0.590] & 0.586 [0.577-0.594] \\
    Competing Random Forest & 0.591 [0.575-0.606] & 0.573 [0.557-0.588] \\
    \noalign{\smallskip}
	\hline
	\noalign{\smallskip}
    SSPN 	& \textbf{0.603 [0.593-0.613]} & \textbf{0.613 [0.598-0.627]}  \\
    \noalign{\smallskip}
	\hline
	\noalign{\smallskip}
    \multicolumn{3}{l}{*p-value $<$ 0.05}
  \end{tabular}
\end{table}
\subsection{SEER} The Surveillance, Epidemiology, and End Results Program (SEER) dataset provides information on breast cancer patients during the years 1992-2007. A total of 72,809 patients experienced breast cancer, cardiovascular disease (CVD), other diseases, or were right-censored. The cohort consists of 23 features, including age, race, gender, morphology information, diagnostic information, therapy information, tumor size, tumor type, etc. Missing values were replaced by mean value for real-valued features and by the mode for categorical features.
1.3\% of the patients experienced CVD and 15.6\% experienced breast cancer. Table \ref{tbl:tbl2} displays the results for this dataset. We notice that for the infrequent adverse event, CVD, the performance gain is negligible while for the frequent breast cancer event, the gain is significant. However, we wish to remind the reader that our focus is on healthcare where even minor gains have the potential to save lives. Considering there are 72,809 patients, a performance improvement even as low as 0.1\% has the potential to save multiple lives and should not be disregarded.
\subsection{Synthetic Data} Due to the relative scarcity of competing risks datasets and methods, we have created an additional synthetic dataset to further validate the performance of our method. We have constructed two stochastic processes with parameters and the event times as follows
\begin{equation}
x_i^1, x_i^2, x_i^3 \sim \mathcal{N}(0, \mathbf{I}),\ T_i^1 \sim \exp\left((x_i^{3})^2+x_i^1 \right),\ T_i^2 \sim \exp\left((x_i^{3})^2+x_i^2 \right)\ .
\end{equation}
where $(x_i^1, x_i^2, x_i^3)$ is the vector of features for patient $i$. For $k = 1,2$, the features $x^k$ only have an effect on the event time for event $k$, while $x^3$ has an effect on the event times of both events. Note that we assume event times are exponentially distributed with a mean parameter depending on both linear and non-linear (quadratic) function of features. 
Given the parameters, we first produced $30,000$ patients; among those, we randomly selected $15,000$ patients (50\%) to be right-censored at a time randomly drawn from the uniform distribution on the interval $[0, \min \{T_i^1, T_i^2 \}]$. (This censoring fraction was chosen to be roughly the same censoring fraction as in the real datasets, and hence to present the same difficulty as found in those datasets.) Table \ref{tbl:tbl3} displays the results for the above dataset. We demonstrate the same consistent performance gain as in the previous case.
\section{Conclusion}
Competing risks settings are pervasive in healthcare. They are encountered in cardiovascular diseases, in cancer, and in the geriatric population suffering from multiple diseases. To solve the challenging problem of learning the model parameters from time-to-event data while handling right censoring, we have developed a novel deep learning architecture for estimating personalized risk scores in the presence of competing risks based on the well-known Siamese network architecture. Our method is able to capture complex non-linear representations missed by classical machine learning and statistical models. Experimental results show that our method is able to outperform existing competing risk methods by successfully learning representations which flexibly describe non-proportional hazard rates with complex interactions between covariates and survival times that are common in many diseases with heterogeneous phenotypes.
\bibliographystyle{splncs04}
\bibliography{icann2018}
\end{document}